\documentclass[11pt]{article}

\usepackage[preprint]{acl}

\usepackage{times}
\usepackage{latexsym}
\usepackage{tcolorbox}
\usepackage{float}

\usepackage[T1]{fontenc}

\usepackage[utf8]{inputenc}

\usepackage{microtype}

\usepackage{inconsolata}

\usepackage{graphicx}
\usepackage{multirow}
\usepackage{color}

\usepackage{subcaption}
\usepackage{tcolorbox}
\tcbuselibrary{xparse}
\newtcolorbox{promptbox}[2][]{fonttitle=\scriptsize, fontupper=\scriptsize, fontlower=\scriptsize, size=title, middle=0mm, before skip balanced=-1mm, #1, title={#2}}
\usepackage{makecell}
\title{Enterprise Benchmarks for Large Language Model Evaluation}

\author{
\\
 \textbf{Bing Zhang\thanks{Equal contribution.}\textsuperscript{1}},
 \textbf{Mikio Takeuchi\footnotemark[1]\textsuperscript{2}},
 \textbf{Ryo Kawahara\footnotemark[1]\textsuperscript{2}},
\\
 \textbf{Shubhi Asthana\textsuperscript{1}},
 \textbf{Md. Maruf Hossain\thanks{Presently Ontario Ministry of Labour, Immigration, Training and Skills Development (views are solely of the author)}\textsuperscript{2}},
 \textbf{Guang-Jie Ren\textsuperscript{1}},
 \textbf{Kate Soule\textsuperscript{3}},
 \textbf{Yada Zhu\thanks{Corresponding author.}\textsuperscript{3}},
\\
 \texttt{bing.zhang@ibm.com}, \texttt{\{mtake, ryokawa\}@jp.ibm.com}, \texttt{w\_maruf@outlook.com}, \\
 \texttt{\{sasthan, gren\}@us.ibm.com}, \texttt{kate.soule@ibm.com}, \texttt{yzhu@us.ibm.com} \\
 \textsuperscript{1}IBM Almaden Research Lab, San Jose, CA, USA \\
 \textsuperscript{2}IBM Research - Tokyo, Japan \\
 \textsuperscript{3}MIT-IBM Watson AI Lab, MA, USA \\
\\
\\
\\}

\begin{document}
\pdfoutput=1
\maketitle

\begin{abstract}
The advancement of large language models (LLMs) has led to a greater challenge of having a rigorous and systematic evaluation of complex tasks performed, especially in enterprise applications. Therefore, LLMs need to be able to benchmark enterprise datasets for various tasks. This work presents a systematic exploration of benchmarking strategies tailored to LLM evaluation, focusing on the utilization of domain-specific datasets and consisting of a variety of NLP tasks. The proposed evaluation framework encompasses 25 publicly available datasets from diverse enterprise domains like financial services, legal, cyber security, and climate and sustainability. 
The diverse performance of 13 models across different enterprise tasks highlights the importance of selecting the right model based on the specific requirements of each task. Code and prompts are available on \href{https://github.com/IBM/helm-enterprise-benchmark}{GitHub}.
\end{abstract}

\section{Introduction}
Large Language Models (LLMs), or foundation models, have garnered significant attention and adoption across various domains due to their remarkable capabilities in natural language understanding and generation. To align with the new era of LLMs, new benchmarks have been proposed recently to probe a diverse set of LLM abilities. For example, BIG-bench (Beyond the Imitation Game benchmark) \cite{srivastava2022beyond} and HELM (Holistic Evaluation of Language Models) \cite{liang2022holistic} attempt to aggregate a wide range of natural language processing (NLP) tasks for holistic evaluation, but often lack domain-specific benchamarks, particularly for enterprise fields such as finance, legal, climate, and cybersecurity. This gap poses challenges for practitioners seeking to assess LLM performance tailored to their needs.

Enterprise datasets, though potentially useful as benchmarks, often face accessibility or regulatory issues. Evaluating LLMs with these datasets can be difficult due to sophisticated concepts or techniques needed to convert use case-based inputs to the standard input format of evaluation harness (e.g., BIG-bench or HELM), which indicates the need for standardized metrics and clear performance benchmarks. This highlights the necessity for robust evaluation frameworks that measure LLM performance in specialized domains.

Emerging enterprise-focused or domain-specific LLMs, such as Snowflake Arctic\footnote{https://www.snowflake.com/blog/arctic-open-efficient-foundation-language-models-snowflake/} and BloombergGPT \cite{wu2023bloomberggpt}, are evaluated with limited enterprise application scope and volume. For textual inputs, Snowflake Arctic is assessed on world knowledge, common sense reasoning, and math. However, academic benchmarks often fail to address the complexities of enterprise applications, such as financial Named Entity Recognition (NER), which requires precise domain language understanding. BloombergGPT is evaluated with several finance datasets, mostly proprietary, and does not include the summarization task.

\begin{figure}
    \centering
    \includegraphics[width=0.98\columnwidth]{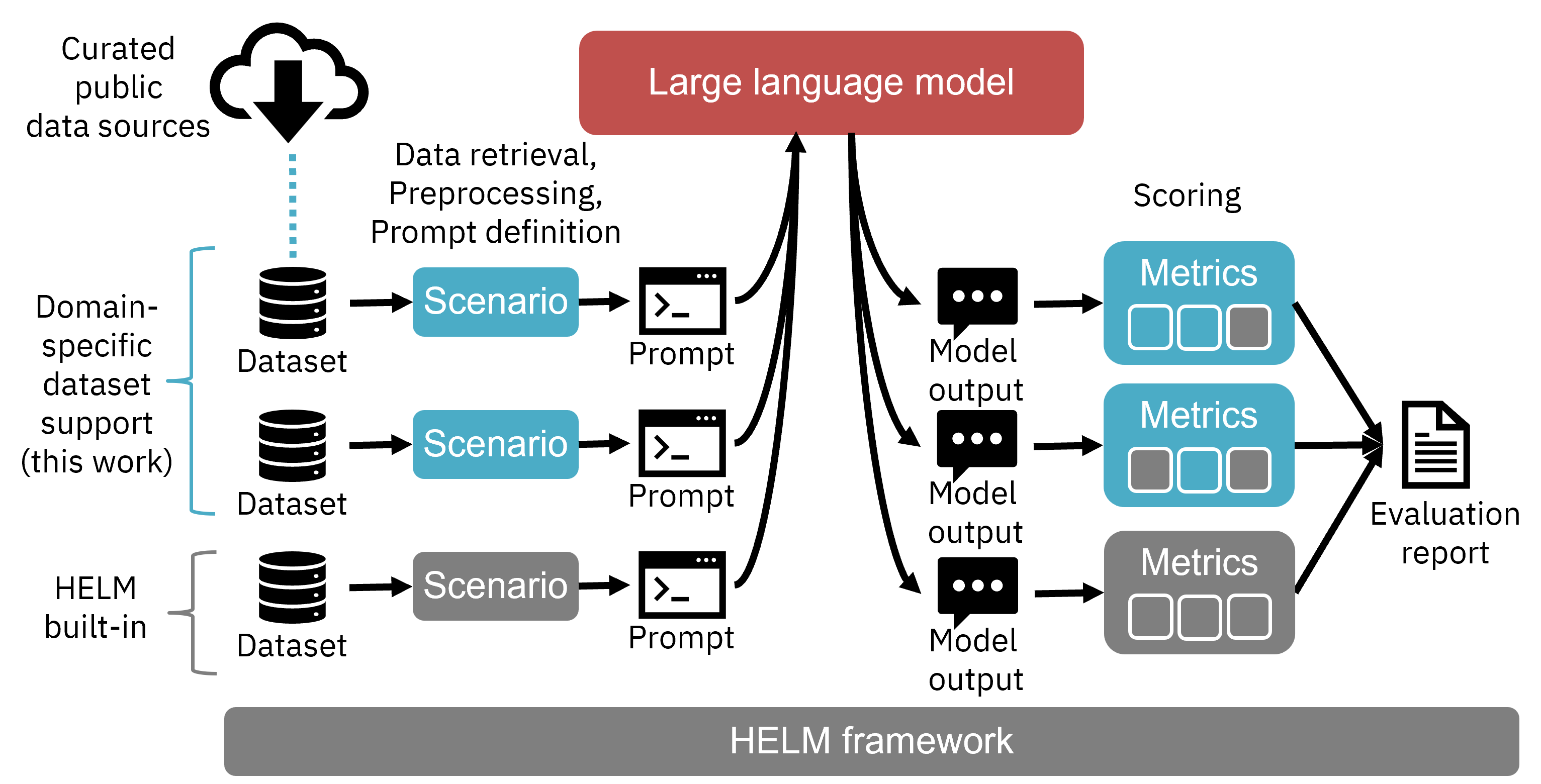}
    \caption{Overview of the enterprise benchmark framework for LLM evaluation.}
    \label{fig:diagram}
\end{figure}

To narrow the gap between LLM development and evaluation in enterprises, we present a framework in Figure \ref{fig:diagram} by augmenting Stanford's HELM that emphasizes the use of enterprise benchmarks that cater specifically to domains such as finance, legal, climate and sustainability, and cyber security. This framework aims to create and adopt standardized benchmarks reflecting real-world application requirements. By enhancing evaluation metrics and providing tailored enterprise-specific prompts, we seek to empower practitioners to assess LLMs more effectively, bridging the gap between LLM development and enterprise needs. This initiative not only addresses the current scarcity of domain-specific evaluation frameworks but also informs better decisions for deploying and optimizing LLM technologies across diverse enterprise environments.

Together, our work makes the following key contributions:
\begin{enumerate}
    \item Bridging gaps for evaluating LLMs in solving domain-specific problems by curating, developing, and implementing a set of benchmarks and performance metrics to the widely adopted HELM evaluation harness.
    \item Enabling researchers and industry practitioners to assess and optimize LLMs tailored to specific domains. 
    \item Evaluate widely used open-source LLMs for domain-specific enterprise applications by conducting extensive experiments with proposed benchmarks and providing prompts for practitioners.
    \item Providing insights and future needs of benchmarking LLMs for enterprise applications.
\end{enumerate}
This paper does not aim to provide an exhaustive evaluation of LLM performance across all enterprise benchmarks; instead, it focuses on specific benchmarks relevant to the study's objectives. Our goal is to highlight the evaluation framework rather than cover every possible enterprise dataset or metric.

In the next section, we delve into the current state-of-the-art LLM evaluation benchmarks. In section 3, we introduce 25 enterprise datasets in 4 domains. Section 4 describes the key design considerations in the development of the benchmark. Experiments and primary results are presented in section 5. The paper concludes in section 6.

\section{Related Work}
Recently, researchers have developed several frameworks to assess various capabilities of LLMs. Examples include HELM \cite{bommasani2023holistic}, MMLU \cite{hendrycks2020measuring}, Big-Bench \cite{lewkowycz2022beyond}, EleutherAI \cite{phang2022eleutherai}, and MMCU \cite{zeng2023measuring}, which are widely used to evaluate LLMs on multiple NLP tasks. Specifically, HELM categorizes potential scenarios and metrics of interest for LLMs. However, these frameworks lack benchmarks and metrics for evaluating LLM performance in enterprise-focused problems. This work leverages the HELM platform, extending its benchmark scenarios and metrics to include task-specific and domain-specific evaluations for LLMs.

Researchers are developing enterprise LLM’s benchmarks in areas such as finance, legal, and sustainability. In FinBen \cite{xie2024finben}, Xie et al. provided finance-based benchmark on 24 financial tasks including information extraction, question answering, risk management, etc. The drawback of this benchmark is that its tasks were designed for the Chinese language, and hence its applicability is limited in American market data and English texts. To counter this, Xu et al.\cite{xu2024superclue} provided fine-granted analysis on diverse financial tasks and applications for six domains and twenty-five specialized tasks in the Chinese language. However, this work cannot be extended to English and other languages and hence is not widely usable.  

Enterprise benchmarks in legal are upcoming with works like Legalbench \cite{guha2024legalbench}, Lawbench \cite{fei2023lawbench}, and LAiW \cite{dai2023laiw}. Lawbench is evaluated on multilingual and Chinese-oriented LLMs while LAiW is Chinese legal LLMs benchmark. Legalbench provides a benchmark on reasoning while the others are evaluating legal foundation inference and complex legal application tasks. 
 
Lastly, in the cybersecurity domain, researchers have contributed to benchmarks like SEvenLLM \cite{ji2024sevenllm}, CyberBench \cite{liucyberbench}, Cyberseceval 2 \cite{bhatt2024cyberseceval} and CyberMetric \cite{tihanyi2024cybermetric}. These benchmarks provide analysis on tasks like cyber advisory and reasoning, question-answering, and cybersecurity incident analysis. Compared to existing benchmarks, our enterprise benchmarks perform sentiment analysis and summarization tasks that have not been tackled in existing art. The benchmarks in our work are open-sourced and consolidated into a widely adopted evaluation framework to enable comprehensive evaluation across reasoning tasks.

\section{Enterprise Benchmarks}
This work introduces benchmark datasets from four specific domains: finance, legal, climate and sustainability, and cyber security, where natural language understanding is crucial for productivity and decision-making. All datasets are curated from open data sources to cover a broad range of natural language tasks and diverse industry use cases within these domains. 
Datasets without reference answers or with fewer than 100 test cases were excluded from the benchmarks.
Although the collected tasks are mostly conventional
(See also Section \ref{sec:limitations}), these tasks are still important in the context of LLM applications.
In addition, the combination of such tasks and domain-specific datasets are still rare and understudied. 
The focus of this paper is 
in catering a means for practitioners to evaluate the performance of processing domain-specific datasets. 
This is because it is known that a general domain LLM might suffer from the degradation of performance when it processes domain-specific data because of the unique terminology and knowledge that are only used in a specific industry. 

As summarized in Table~\ref{tab: fin-data}, the finance benchmarks include 10 datasets collected from important use cases such as market prediction based on earnings call transcripts, entity recognition for retrieving information from U.S. Securities and Exchange Commission (SEC) filings\footnote{https://www.sec.gov/}, and understanding news and reports. The tasks range from classification and NER to QA and long document summarization. NER is crucial for many applications in digital finance, and numerical NER is a particularly challenging task for language models. ConvFinQA provides multi-turn conversational financial QA data involving information extraction from tables and numerical reasoning, offering a critical lens for evaluating LLMs' numerical reasoning capabilities.

\begin{table*}[htb]
\scriptsize
\caption{Finance Benchmarks Overview}
\label{tab: fin-data}
    \centering
\begin{tabular}{|p{0.1\textwidth}|p{0.06\textwidth}|p{0.1\textwidth}|p{0.45\textwidth}|p{0.05\textwidth}|p{0.06\textwidth}|}
\hline 
Task & Task Description & Dataset & Dataset Description & N-shot Prompt &  Metric\tabularnewline
\hline 
\hline 
\multirow{2}{0.2\columnwidth}{Classification} & 2 Classes  & Earnings Call Transcripts~\cite{Roozen2021} & Earnings call transcripts, the related stock prices and the sector index in terms of volume & 5-shot & Weighted F1\tabularnewline
\cline{2-6} 
& 9 Classes & News Headline~\cite{sinha2020impact}&  The gold commodity news annotated into various dimensions & 5-shot & Weighted F1\tabularnewline
\hline 
\multirow{3}{0.2\columnwidth}{Named Entity \\
Recognition} & 4 numerical entities & Credit Risk Assessment (NER)~\cite{salinas-alvarado-etal-2015-domain} & Eight financial agreements (totalling 54,256 words) from SEC filings were manually annotated for entity types: location, organization person and miscellaneous & 20-shot & Entity F1 \tabularnewline
\cline{2-6} 
 & 4522 Numerical Entities & KPI-Edgar~\cite{Deu_er_2022} & A dataset for Joint NER and Relation Extraction building on financial reports uploaded to the Electronic Data Gathering, Analysis, and Retrieval (EDGAR) system, where the main objective is to extract Key Performance Indicators (KPIs) from financial documents and link them to their numerical values and other attributes & 20-shot & Adj F1\tabularnewline
\cline{2-6}  
 & 139 Numerical Entities & FiNER-139~\cite{loukas-etal-2022-finer} & 1.1M sentences annotated with extensive Business Reporting Language (XBRL) tags extracted from annual and quarterly reports of publicly-traded companies in the US, focusing on numeric tokens, with the correct tag depending mostly on context, not the token itself.  & 10-shot & Entity F1\tabularnewline
\hline 
\multirow{4}{0.2\columnwidth}{Question \\
Answering} & Document relevance ranking & Opinion-based QA (FiQA)~\cite{FinQA2018} & Text documents from different financial data sources (microblogs, reports, news) for ranking document relevance based on opinionated questions, targeting mined opinions and their respective entities, aspects, sentiment polarity and opinion holder.  & 5-shot & RR@10\tabularnewline
\cline{2-6}
 & 3 Classes & Sentiment Analysis (FiQA SA)~\cite{FinQA2018} & Text instances in the financial domain (microblog message, news statement or headline) for detecting the target aspects which are mentioned in the text (from a pre-defined list of aspect classes) and predict the sentiment score for each of the mentioned targets.  & 5-shot  & Weighted F1\tabularnewline
\cline{2-6}
 & Ranking & Insurance QA~\cite{feng2015applying} & Questions from real world users and answers with high quality composed by professionals with deep domain knowledge collected from the website Insurance Library~\footnote{https://www.insurancelibrary.com/} & 5-shot & RR@10\tabularnewline
\cline{2-6}
 & Exact Value Match & Chain of Numeric Reasoning (ConvFinQA)~\cite{chen2022convfinqa} & Multi-turn conversational finance question answering data for exploring the chain of numerical reasoning & 1-shot & Accuracy \tabularnewline
\hline 
Summarization & Long Documents & Financial Text Summarization (EDT)~\cite{zhou2021trade} & 303893 news articles ranging from March 2020 to May 2021 for abstractive text summarization  & 5-shot & Rouge-L\tabularnewline
\hline 
\end{tabular}
\end{table*}

Similarly, the seven legal benchmarks in Appendix/Table~\ref{tab: legal-data} contain rich NLP tasks and important use cases, such as legal opinion classification, legal judgment prediction, and legal contract summarization. Climate and sustainability is an emerging domain for LLM applications, including summarizing claims and understanding human concerns like wildfires and climate change. Due to limited open-source datasets with quality labels, three benchmarks are curated, as shown in Appendix/Table~\ref{tab: climate-data}. Security-related tasks, including classification and summarization of textual documents such as network protocol specifications, malware reports, vulnerability, and threat reports, are curated and shown in Appendix/Table~\ref{tab: cyber-data}.

\section{Benchmark Development}

Recent LLMs, primarily based on the decoder-only Transformer architecture, have unique capabilities and limitations, such as in-context learning (few-shot learning) and input token length constraints. Domain-specific benchmark datasets are often designed for conventional machine learning models or different architectures (such as BERT), necessitating adaptations in datasets and task implementations.

In HELM, a \textit{scenario} represents an evaluation task with a specific dataset and corresponding metrics. These adaptations are incorporated into the development of the scenarios.

\subsection{Classification Task} \label{sec:scenario-design-classification}

In a classification task, a model is asked to generate the name of a class of the input sample directly as an output\footnote{A perplexity/log-likelihood-based approach is not used for easier comparison with the existing HELM scenarios that generate labels directly.}. 
It is better to use natural language words as the class names (e.g., positive/neutral/negative) than to use symbolic names (see the discussion in Section \ref{sec:scenario-design-ner}). One usually needs to provide few-shot examples to ensure that a model does not generate tokens other than the class names.

For multi-class classification tasks with more than 20 classes, defining all classes in a prompt and covering them in in-context learning examples is challenging due to input token length limits. This work simplifies the task by selecting the top-$k$ classes based on their distributions, where $k$ is typically less than 10.

In addition to HELM's built-in micro- and macro-F1 scores, the Weighted F1 score as implemented in~\cite{scikit-learn-2024} is added as a performance metric.

\subsection{Named Entity Recognition Task} \label{sec:scenario-design-ner}
A conventional NER task is formalized as a sequence-to-sequence task, where the input is a sequence of tokens. A system classifies whether each token is a part of a named entity and identifies its category (e.g., person, location, organization, etc.). 
Then the system generates a sequence of corresponding tags (so-called BIO tags) 
in the same order as the input tokens\cite{Cui2021template}. 
However, in our preliminary experiments, this approach did not work well with LLMs. This seems to be because BIO tags are unknown to pre-trained LLMs.
In addition, few-shot examples consume many tokens if the inputs and the tags are provided in a seq-to-seq manner. 
The former issue can be overcome by formalizing the task as a
translation to an augmented natural language\cite{Paolini2021}, but the latter issue still exists.

Due to these difficulties, a simplified approach used in the evaluation of BloombergGPT\cite{wu2023bloomberggpt} for NER tasks is employed in this work.
Instead of generating a sequence of BIO tags, a model extracts only named entities and their categories, and answers those in a list of paired natural language phrases
(e.g., New York (location), etc.).
There is another report that such use of natural language words
improves NER performance in a low-resource domain because of a transfer of pre-trained knowledge from the general language domain\cite{Cui2021template}. In some scenarios, the number of categories is reduced, as explained in the previous section.
Even with these simplifications, a NER task requires more in-context learning examples than a classification task. 

To support the above task of extracting named entities, a new metric called Entity F1 is added.
For each test sample, predicted named entities and the categories of those are
compared with those in the ground-truth, to compute 
true positives, false positives, and false negatives. Those are aggregated population-wide to compute
the precision, recall, and F1 scores.

\subsection{Question and Answering Task} \label{sec:scenario-design-qa}
There are several types of QA tasks, some of which overlap with information retrieval tasks. In many business applications, one is requested to answer a question
based on a given set of documents (e.g., product manuals, FAQs, medical papers, regulations, etc.).
This involves a ranking of answer candidates with respect to their relevance to the user's question.
However, LLMs struggle with these operations because handling multiple answer candidates in a single prompt consumes many tokens.

Alternatively, the "point-wise" approach provided in HELM is adopted\cite{liang2022holistic}. 
For a question $q_{i}$, there are $k$ pre-defined answer candidates $\{a_{ij}|j=1,\cdots k\}$
and one prompts the following question to a model: "Does $a_{ij}$ answer the question $q_{i}$? Answer in yes or no."
From this prompt, one can obtain a pair of the output text $b_{ij} \in \{\mbox{"yes"}, \mbox{"no"}\}$ and its log probability $c_{ij}$ from the model.
Note that the log probability of a token 
is available in APIs of the most of public LLM services,
and plays a role of the confidence of the output.
As a result, one collects $k$ output pairs $\{(b_{ij}, c_{ij})\}$ for one question $q_{i}$.
Then, one can rate an answer candidate that has $b_{ij}=\mbox{"yes"}$ with higher $c_{ij}$
to be higher and that has $b_{ij}=\mbox{"no"}$ with higher $c_{ij}$ to be lower, which defines the ranking.

\subsection{Summarization Task} \label{sec:scenario-design-summarization}

In a summarization task, one needs to handle a long document as an input. Therefore, the input token length limit becomes a severe issue.
In this study, this issue is handled by selecting relatively shorter samples and truncating the end of the samples to preserve the original context as much as possible.

In this study, conventional ROUGE scores are used as performance metrics.

\section{Experiments and Results}
\label{sec:results}
This evaluation is conducted by augmenting HELM's framework to encompass $25$ publicly available task datasets from multiple domains, namely financial, legal, climate and cyber security. 
For each benchmark, the evaluation is done on a specific configuration. 
The intention of this section is 
to demonstrate the usefulness for practitioners of our benchmarks in evaluating candidate models with their own settings.

\subsection{Evaluated Models} \label{sec:evaluation-models}
Here, the evaluation models are selected from the best-performing open-sourced models under 70 billion parameters based on model size, type of training data, accessibility, and model tuning method. 
Specifically 1) \textbf{LLaMA2}~\cite{metaAI-llama} is a series of advanced LLM ranging from 7 billion to 70 billion parameters. It is designed to perform well across various tasks and is notable for its versatility and robustness in natural language understanding and generation. 2) \textbf{GPT-NeoX-20B}~\cite{black2022gptneox20b} 
is particularly known for its impressive performance in text generation tasks, providing a strong alternative to proprietary models due to its open-source nature and large model size. 3) \textbf{FLAN-UL2}~\cite{tay2023ul2} is another state-of-the-art model that has been fine-tuned with a focus on instruction understanding and task completion. It excels in few-shot and zero-shot learning scenarios, making it highly effective for various practical applications. 
4) \textbf{Phi-3.5}~\cite{abdin2024phi3technicalreporthighly} is a family of powerful, small language models (SLMs) with groundbreaking performance at low cost and low latency. 5) \textbf{Mistral 7B}~\cite{jiang2023mistral7b} is a series of 7-billion-parameter language models engineered for superior performance and efficiency.
6) The \textbf{granite.13b} models\footnote{https://ibm.biz/techpaper-granite-13b} are set of the latest open-sourced enterprise-focused models. The datasets used in this second tranche of training tokens includes some finance and legal datasets, such as FDIC, Finance Text Books, EDGAR Filings, etc. The base model serves as the foundation for two variants: granite.13b.instruct.v2 and granite.13b.chat.v2.1. Granite.13b.instruct has undergone supervised fine-tuning to enable better instruction following \cite{wei2021finetuned}, making it suitable for completing enterprise tasks via prompt engineering. Granite.13b.chat uses novel alignment methods to improve generation quality, reduce harm, and ensure outputs are helpful and follow social norms \cite{conover2023free, bai2022training, kim2022prosocialdialog}.

All the 13 models are evaluated in our benchmarks, regardless of the purposes of the models (i.e., for chat, etc.). As we will see in the following sections, the relation between the performance of a task and the intended purpose of a model is not straightforward.

\subsection{Evaluation Setup}
In this study, the data source-provided train and test splits are used whenever possible. Model performance is reported based on test or validation examples, depending on the availability of test labels. If the train and test splits do not exist, a task-specific ratio of the data is selected as the test split, with the remainder used as the train split. In-context learning examples are sampled from the train split. The number of few-shot examples provided to the model varies by task and is detailed in Appendix/Tables~\ref{tab: fin-data}, \ref{tab: legal-data}, \ref{tab: climate-data}, and \ref{tab: cyber-data}. Note that, by default in HELM, only one set of randomly sampled examples is used across all test cases of a given benchmark. If the training context examples are unfortunately not good, the performance of all the models will be affected. A contextual sampling strategy based on nearness to the test sample through some similarity measure is future work.

For in-context learning, this work adopts HELM's sampling strategy, which includes samples from minority classes.  This is different from a conventional uniformly random sampling, where samples in a minority class tend to be ignored in the case of a few-shot sampling. This sampling strategy is especially important when the distribution of the labels is imbalanced and its application needs a capability to detect the minority class samples (e.g., anomaly detection). 
However, this strategy has less impact when the focus is on accurately classifying majority class samples. 

For the current evaluation, all the models use the same parameters and the same context examples. Standard prompts (see the techniques of few-shot-prompting and zero-shot-prompting and examples of prompts\footnote{https://www.promptingguide.ai/techniques/fewshot}), without chain-of-thought prompting~\cite{wei2023chainofthought}, or system prompts are used. For News Headline and FiQA SA, the prompts are taken from BloombergGPT~\cite{wu2023bloomberggpt}. The prompts for each scenario are shown in Appendix \ref{sec:appendix-prompts}. To ensure reproducibility, a fixed random seed and the greedy decoding method without repetition penalty are used. After generating the output, standard text normalization (i.e., moving articles, extra white spaces, and punctuations followed by lowering cases) is applied before matching texts. In ConvFinQA, a regular expression is used to match the floating-point numbers.

\subsection{Evaluation Results}
\label{sec:results_finance}

\textbf{Finance Benchmark} The results presented in Table~\ref{tab: fin-results} provide the evaluation results of 13 models across a range of financial NLP tasks, including classification, NER, QA, and summarization. Each task was assessed using the best-fitted metrics to determine the performance of different models.

For classification tasks, the highest Weighted F1 score was achieved by the granite.13b.instruct.v2 model in the Earnings Call Transcripts classification, demonstrating its strong performance in extracting relevant information from earnings calls. For News Headline classification, the llama2.70b.chat model achieved the highest Weighted F1 score, indicating its effectiveness in handling short text classification tasks.

NER was evaluated using three different tasks: Credit Risk Assessment, KPI-Edgar, and FiNER-139. The llama2.70b.chat model outperformed others in Credit Risk Assessment with the highest Entity F1 score. In KPI-Edgar, the llama2.70b model achieved the best Modified Adjusted F1 (Adj F1) score, while gpt-neox-20b led in FiNER-139 with the highest Entity F1 score, showcasing its capability in identifying financial entities accurately.

Among the diverse QA tasks, the flan-ul2 model excelled in FiQA-Opinion and Insurance QA with the highest RR scores, highlighting its proficiency in answering complex questions with limited context. For FiQA SA, the llama2.70b.chat model obtained the highest Weighted F1 score. The granite.13b.chat.v2.1 model stood out in ConvFinQA with the highest accuracy, indicating its robustness in handling multi-turn financial QA tasks involving numerical reasoning.

For Text Summarization, the flan-ul2 model achieved the highest Rouge-L score, demonstrating its ability to generate concise and relevant summaries from financial texts. Compared to other domains, the financial benchmark evaluation extensively encompasses all kinds of LLM tasks.


\begin{table*}
\scriptsize
\caption{Finance Benchmark Evaluation Results per Task.}
\label{tab: fin-results}
    \centering
\begin{tabular}{|p{0.15\textwidth}|p{0.06\textwidth}|p{0.05\textwidth}|p{0.055\textwidth}|p{0.04\textwidth}|p{0.045\textwidth}|p{0.05\textwidth}|p{0.045\textwidth}|p{0.05\textwidth}|p{0.05\textwidth}|p{0.07\textwidth}|}

\hline 
Task &\multicolumn{2}{|c|}{Classification} & \multicolumn{3}{c|}{Named Entity Recognition} & \multicolumn{4}{c|}{Question Answering} & \multicolumn{1}{c|}{Summarization}\tabularnewline
\hline
 & Earnings Call Transcripts & News Headline & Credit Risk Assessment\ & KPI-Edgar & FiNER-139 & FiQA-Opinion & Insurance QA & FiQA SA & ConvFinQA & Financial Text Summarization\tabularnewline
\hline 
\hline 
Metrics & Weighted F1 & Weighted F1 & Entity F1 & Adj F1 & Entity F1 & RR @10 & RR@5 & Weighted F1 & Accuracy & Rouge-L\tabularnewline
\hline
phi-3-5-mini-instruct & 0.411 & 0.800 & 0.417 & 0.421 & 0.677 & 0.605 & 0.350 & 0.824 & 0.277 & 0.368\tabularnewline
\hline
mistral-7b-instruct-v0-3 & 0.453 & 0.794 & 0.396 & 0.588 & 0.686 & 0.569 & 0.414 & 0.838 & 0.280 & 0.390\tabularnewline
\hline
llama2.7b & 0.411 & 0.753 & 0.427 & 0.419 & 0.661 & 0.599 & 0.238 & 0.744 & 0.233 & 0.154\tabularnewline
\hline 
llama2.7b.chat & 0.511 & 0.829 & 0.463 & 0.451 & 0.627 & 0.557 & 0.505 & 0.693 & 0.198 & 0.422\tabularnewline
\hline 
llama2.13b & 0.411 & 0.584 & 0.483 & 0.463 & 0.689 & 0.660 & 0.546 & 0.800 & 0.260 & 0.337\tabularnewline
\hline 
llama2.13b.chat & 0.541 & 0.744 & 0.425 & 0.539 & 0.672 & 0.667 & 0.425 & 0.849 & 0.261 & 0.420\tabularnewline
\hline 
llama2.70b & 0.411& 0.818 & 0.373 & \bf{0.714} & 0.715 & 0.769 & 0.477 & 0.837 & 0.344 & 0.398\tabularnewline
\hline 
llama2.70b.chat & 0.504 & \bf{0.840} & \bf{0.550} & 0.679 & 0.694 & 0.638 & 0.472 & \bf{0.859} & 0.304 & 0.427\tabularnewline
\hline 
granite.13b.v2 (base) & 0.411 & 0.804 & 0.416 & 0.368 & 0.739 & 0.606 & 0.183 & 0.798 & 0.368 & 0.374\tabularnewline
\hline 
granite.13b.instruct.v2 & \bf{0.618} & 0.817 & 0.411 & 0.295 & 0.680 & 0.669 & 0.605 & 0.776 & 0.386 & 0.398\tabularnewline
\hline 
granite.13b.chat.v2.1 & 0.411 & 0.808 & 0.475 & 0.504 & 0.765 & 0.584 & 0.613 & 0.795 & \bf{0.407} & 0.416\tabularnewline
\hline 
gpt-neox-20b & 0.411 & 0.630 & 0.351 & 0.308 & \bf{0.774} & 0.503 & 0.197 & 0.771 & 0.266 & 0.176\tabularnewline
\hline 
flan-ul2 & 0.411 & 0.829 & 0.259 & 0.011 & 0.446 & \bf{0.804} & \bf{0.723} & 0.811 & 0.254 & \bf{0.428}\tabularnewline
\hline 
\end{tabular}
\end{table*}%



\label{sec:results_legal_climate}

\textbf{Legal Benchmark}
The results in Appendix/Table~\ref{tab: legal-results} highlight the performance of various models across legal tasks. For classification, the mistral-7b-instruct-v0-3 model excelled in Legal Sentiment Analysis (0.727 Weighted F1) and UNFAIR-ToS (0.720 Weighted F1), while granite.13b.instruct.v2 led in Legal Judgement Prediction (0.863 Weighted F1). In QA, granite.13b.instruct.v2 achieved the highest F1 score (0.790) in the CaseHOLD task. For summarization, granite.13b.instruct.v2 was best in BillSum (0.422 Rouge-L), and llama2.70b topped Legal Summarization (0.291 Rouge-L).

\textbf{Climate and Sustainability Benchmark}
Appendix/Table~\ref{tab: climate-results} shows the evaluation of models on climate and sustainability tasks. The flan-ul2 model performed best in Reddit Climate Change classification (0.560 Weighted F1) and SUMO Climate Claims summarization (0.258 Rouge-L), while the phi-3-5-mini-instruct model led in Wildfires and Climate Change Tweets classification (0.796 Weighted F1).

\textbf{Cyber Security Benchmark}
Appendix/Table~\ref{tab: cyber-results} presents the performance of models on cyber security tasks. The phi-3-5-mini-instruct model excelled in SPEC5G classification (0.527 Weighted F1), while llama2.70b.chat led in CTI-to-MITRE with NLP (0.812 F1). For SecureNLP and IoTSpotter, flan-ul2 and llama2.70b achieved the highest Binary F1 scores (0.369 and 0.905, respectively). In summarization, gpt-neox-20b was the best in SPEC5G Summarization (0.453 Rouge-L).

Across all domains, the results indicate that different models excel in various tasks depending on their training process and architecture. In the legal domain, the granite.13b.instruct.v2 and llama2.70b.chat models showed strong performance in classification and question answering tasks, respectively. For climate and sustainability, the flan-ul2 model demonstrated its effectiveness in both classification and summarization. In cyber security, the llama2.70b and flan-ul2 models exhibited superior performance across multiple classification tasks.

It is easy to expect that a larger model performs better than a smaller model.
However, inference cost is an important factor in choosing an LLM, and is correlated to its parameter size. For example, both the granite.13b.instruct.v2 and llama2.70b.chat models works well in Legal classification and QA tasks. However, if one considers the parameter size of those models, 
granite.13b.instruct.v2 can be said to be more cost-effective. This could be attributed to its use of
legal text datasets for pre-training (Section \ref{sec:evaluation-models}). 
However, such data source information is usually not available even in the case of an open source model.
Thus, it is difficult to predict the performance of a model without benchmarks, especially in the case
of a domain-specific performance of a small model.

These evaluations underscore the importance of selecting the appropriate model based on the specific requirements and nature of the task at hand. The diversity in performance also highlights the potential for further model optimization and specialization in these domains.

\section{Conclusion and Future Work} \label{sec:conclusion}
In summary, this work advances the evaluation of LLMs in domain-specific contexts by consolidating benchmark datasets and incorporating unique performance metrics into Stanford's HELM framework. This enables researchers and industry practitioners to assess and optimize LLMs for specific domains. We demonstrated the effectiveness of these evaluations on widely used 13 LLMs through extensive experiments on 25 publicly available benchmarks in financial, legal, climate, and cyber security domains, providing practical prompts for practitioners. Our analysis offers valuable insights and highlights future needs for benchmarking LLMs in specialized applications. In addition to the code and prompts being open-sourced, the Stanford HELM team is incubating the benchmarks and code into HELM to facilitate community adoption of this work. 

\subsection{Limitations} \label{sec:limitations}
In this study, the same prompts are used for evaluating all models and scenarios. Recent LLMs suggest model-specific prompts with meta-level information. However, HELM's model-agnostic architecture complicates applying different prompts to different models. Future work will explore emerging applications such as Retrieval-Augmented-Generation (RAG) \cite{Yu2024} and Chain-of-Thought (CoT) question answering \cite{Suzgun2023} to optimize prompts.
Evaluation of different aspects of industry needs on LLMs such as instruction following, multi-lingual capabilities including machine translation, and a case study are also included in future work.

\section{Acknowledgments}
This work is funded by IBM Research and MIT-IBM Watson AI Lab. We would like to thank David Cox, Rameswar Panda, Maruf Hossain, Naoto Satoh, Futoshi Iwama and Alisa Arno for their guidance and constructive comments. The views and conclusions are those of the authors and should not be interpreted as representing that of IBM or the government.

\bibliography{acl_main}

\appendix

\section{Appendix}
\label{sec:appendix}

\subsection{Benchmarks Overview}
\label{sec:benchmarks-overview}
The data tables summarize key benchmarking information. Each table includes the \textit{Task}, which specifies the problem, and the \textit{Task Description}, explaining its nature. The \textit{Dataset} column names the data used, with the \textit{Dataset Description} detailing its characteristics. The \textit{N-shot Prompt} indicates the number of examples used for few-shot learning, which is provided to ensure consistency and replicability. Lastly, the \textit{Metric} column outlines the evaluation metrics used to measure model performance. Table \ref{tab: fin-data}, and \ref{tab: legal-data} to \ref{tab: cyber-data} present the overview of benchmarks in the domain of finance, legal, climate and sustainability, and cybersecurity, respectively.

\begin{table*}[htb]
\scriptsize
\caption{Legal Benchmarks Overview}
\label{tab: legal-data}
    \centering
\begin{tabular}{|p{0.1\textwidth}|p{0.06\textwidth}|p{0.1\textwidth}|p{0.45\textwidth}|p{0.05\textwidth}|p{0.06\textwidth}|}
\hline 
Task & Task Description & Dataset & Dataset Description & N-shot Prompt &  Metric\tabularnewline
\hline 
\hline 
\multirow{2}{0.2\columnwidth}{
Classification} & 3 Classes & Legal Sentiment Analysis~\footnotemark{} & Legal opinion categorised by sentiment & 5-shot & Weighted F1\tabularnewline
\cline{2-6} 
& Multi-classes & UNFAIR-ToS~\cite{lippi2019claudette} & 
The UNFAIR-ToS dataset contains 50 Terms of Service (ToS) from online platforms. The dataset has been annotated on the sentence-level with 8 types of unfair contractual terms, meaning terms (sentences) that potentially violate user rights according to EU consumer law. & 5-shot & Weighted F1\tabularnewline
\cline{2-6} 
& 2 Classes  & Legal Judgement Prediction~\cite{chalkidis-etal-2019-neural} & Legal judgment prediction is the task of automatically predicting the outcome of a court case, given a text describing the case’s facts. This English legal judgment prediction dataset contains cases from the European Court of Human Rights. & 5-shot & Weighted F1\tabularnewline
\hline 
\multirow{4}{0.2\columnwidth}{Question \\
Answering} & Multi-choice QA & CaseHOLD~\cite{zhengguha2021} & The CaseHOLD dataset (Case Holdings On Legal Decisions) provides 53,000+ multiple choice questions with prompts from a judicial decision and multiple potential holdings, one of which is correct, that could be cited. 
& 2-shot & F1\tabularnewline
\hline 
Summarization & Summarization of US Legislations & BillSum~\cite{Eidelman_2019} & The BillSum dataset consists of three parts: US training bills, US test bills and California test bills. The US bills were collected from the Govinfo service provided by the United States Government Publishing Office (GPO). 
For California, bills from the 2015-2016 session were scraped directly from the legislature's website; the summaries were written by their Legislative Counsel.  & 0-shot & Rouge-L\tabularnewline
\cline{2-6}  
& Contract Summarization & Legal Summarization~\cite{manor-li-2019-plain} &  Legal text snippets paired with summaries written in plain English. The summaries involve heavy abstraction, compression, and simplification.  & 0-shot & Rouge-L\tabularnewline
\hline 
\end{tabular}
\end{table*}

\begin{table*}[htb]
\scriptsize
\caption{Climate and Sustainability Benchmarks Overview}
\label{tab: climate-data}
    \centering
\begin{tabular}{|p{0.1\textwidth}|p{0.06\textwidth}|p{0.1\textwidth}|p{0.45\textwidth}|p{0.05\textwidth}|p{0.06\textwidth}|}
\hline 
Task & Task Description & Dataset & Dataset Description & N-shot Prompt &  Metric\tabularnewline
\hline 
\hline 
\multirow{2}{0.2\columnwidth}{Classification} & 2 Classes & Reddit Climate Change~\footnotemark{} & All the mentions of climate change on Reddit before Sep 1 2022. & 5-shot & Weighted F1\tabularnewline
\cline{2-6} 
 & 2 Classes  & Wildfires and Climate Change Tweets~\footnotemark{} & Tweets during the peach of the wildfire season in late summer and early fall of 2020 from public and government agencies. & 5-shot & Weighted F1\tabularnewline
\hline 
Summarization & Generating Fact Checking Summaries & SUMO Climate Claims~\cite{mishra-etal-2020-generating} & Climate claims from news or webs & 0-shot & Rouge-L\tabularnewline
\hline 
\end{tabular}
\end{table*} 

\begin{table*}[htb]
\scriptsize
\caption{Cyber Security Benchmarks Overview}
\label{tab: cyber-data}
    \centering
\begin{tabular}{|p{0.1\textwidth}|p{0.06\textwidth}|p{0.1\textwidth}|p{0.45\textwidth}|p{0.05\textwidth}|p{0.06\textwidth}|}
\hline 
Task & Task Description & Dataset & Dataset Description & N-shot Prompt &  Metric\tabularnewline
\hline 
\hline 
\multirow{2}{0.2\columnwidth}{Classification} & 3 Classes & SPEC5G~\cite{karim2023spec5g} & SPEC5G is a dataset for the analysis of natural language specification of 5G Cellular network protocol specification. SPEC5G contains 3,547,587 sentences with 134M words, from 13094 cellular network specifications and 13 online websites. It is designed for security-related text classification and summarisation. & 5-shot & Weighted F1\tabularnewline
\cline{2-6} 
 &  Multi-
classes  & CTI-to-MITRE with NLP~\cite{orbinato2022automatic} & This dataset contains samples of CTI (Cyber Threat Intelligence) data in natural language, labeled with the corresponding adversarial techniques from the MITRE ATT\&CK framework. & 10-shot  & F1\tabularnewline
\cline{2-6} 
& Multi-
classes & TRAM~\footnotemark{} & The Threat Report ATT\&CK Mapper dataset contain sentences from CTI reports labeled with the ATT\&CK techniques  & 20-shot & Macro F1\tabularnewline
\cline{2-6} 
&  2 Classes  & SecureNLP~\cite{phandi2018semeval} & Semantic Extraction from CybersecUrity REports using Natural Language Processing (SecureNLP), a dataset on annotated malware report. & 5-shot & Binary F1\tabularnewline
\cline{2-6} 
&  2 Classes  & IoTSpotter~\cite{jin2022understanding} & The IoTSpotter dataset is a collection of corpus and IoTSpotter identification results related to Internet of Things (IoT) devices and their security vulnerabilities. & 14-shot & Binary F1\tabularnewline
\hline 
Summarization & Text to Summary & SPEC5G~\cite{karim2023spec5g} & \textit{The same as above.} & 0-shot & Rouge-L\tabularnewline
\hline 
\end{tabular}
\end{table*}

\subsection{Evaluation Results}
\label{sec:aditional-eval-results}
Table \ref{tab: legal-results} to \ref{tab: cyber-results} shows the LLMs evaluation results of legal, climate and cyber security benchmarks. We have discussed these results in section\ref{sec:results_legal_climate}.

\begin{table*}
\scriptsize
\caption{Legal Benchmark Evaluation Results per Task.}
\label{tab: legal-results}
    \centering
\begin{tabular}{|p{0.14\textwidth}|p{0.07\textwidth}|p{0.07\textwidth}|p{0.07\textwidth}|p{0.07\textwidth}|p{0.07\textwidth}|p{0.07\textwidth}|p{0.07\textwidth}|}
\hline 
Task &\multicolumn{3}{|c|}{Classification} & \multicolumn{1}{c|}{Question Answering} & \multicolumn{2}{c|}{Summarization}\tabularnewline
\hline 
 & Legal Sentiment Analysis  & UNFAIR-ToS  & Legal Judgement Prediction\ & CaseHOLD & BillSum & Legal Summarization\tabularnewline
\hline 
\hline 
Metrics & Weighted F1 & Weighted F1 & Weighted F1 & F1 & Rouge-L & Rouge-L\tabularnewline
\hline
phi-3-5-mini-instruct & 0.594 & 0.464 & 0.739 & 0.767 & 0.311 & 0.205\tabularnewline
\hline
mistral-7b-instruct-v0-3 & \bf{0.727} & \bf{0.720} & 0.845 & 0.696 & 0.312 & 0.255\tabularnewline
\hline 
llama2.7b & 0.604 & 0.00003 & 0.122 & 0.583 & 0.269 & 0.258\tabularnewline
\hline 
llama2.7b.chat & 0.590 & 0.557 & 0.819 & 0.467 & 0.313 & 0.235\tabularnewline
\hline 
llama2.13b & 0.456 & 0.003 & 0.273 & 0.683 & 0.267 & 0.261\tabularnewline
\hline 
llama2.13b.chat & 0.699 & 0.340 & 0.800 & 0.727 & 0.328 &  0.246\tabularnewline
\hline 
llama2.70b & 0.566 & 0.120 & 0.819 & 0.760 & 0.038 & \bf{0.291}\tabularnewline
\hline 
llama2.70b.chat & 0.713 & 0.698 & 0.827  & 0.720 & 0.339 & 0.272\tabularnewline
\hline 
granite.13b.v2 (base) & 0.394 & 0.0003 & 0.835  & 0.524 & 0.261 & 0.255\tabularnewline
\hline 
granite.13b.instruct.v2 & 0.620 & 0.428 & \bf{0.863} & \bf{0.790} & \bf{0.422} & 0.193\tabularnewline
\hline 
granite.13b.chat.v2.1 & 0.614 & 0.092 & 0.806 & 0.688 & 0.302 & 0.223\tabularnewline
\hline 
gpt-neox-20b & 0.211 & 0.121 & 0.695 & 0.503 & 0.249 & 0.223\tabularnewline
\hline 
flan-ul2 & 0.646 & 0.302 & 0.073 & 0.780 & 0.234 & 0.173\tabularnewline
\hline
\end{tabular}
\end{table*}

\begin{table*}
\scriptsize
\caption{Climate and Sustainability Benchmark Evaluation Results per Task.}
\label{tab: climate-results}
    \centering
\begin{tabular}{|p{0.15\textwidth}|p{0.08\textwidth}|p{0.08\textwidth}|p{0.07\textwidth}|}
\hline 
Task &\multicolumn{2}{|c|}{Classification} & \multicolumn{1}{c|}{Summarization}\tabularnewline
\hline 
 & Reddit Climate Change & Wildfires and Climate Change Tweets & SUMO Climate Claims\tabularnewline
\hline 
\hline 
Metrics & Weighted F1 & Weighted F1 & Rouge-L \tabularnewline
\hline
phi-3-5-mini-instruct & 0.470 & \bf{0.796} & 0.190\tabularnewline
\hline
mistral-7b-instruct-v0-3 & 0.457 & 0.761 & 0.210\tabularnewline
\hline 
llama2.7b & 0.364 & 0.549 & 0.048\tabularnewline
\hline 
llama2.7b.chat & 0.425 & 0.783 & 0.156\tabularnewline
\hline 
llama2.13b & 0.414 & 0.686 & 0.093\tabularnewline
\hline 
llama2.13b.chat & 0.460 & 0.699 & 0.156\tabularnewline
\hline 
llama2.70b & 0.458 & 0.688 & 0.126\tabularnewline
\hline llama2.70b.chat & 0.498 & 0.748 & 0.194\tabularnewline
\hline 
granite.13b.v2 (base) & 0.315 & 0.722 & 0.022\tabularnewline
\hline 
granite.13b.instruct.v2 & 0.523 & 0.646 & 0.199\tabularnewline
\hline 
granite.13b.chat.v2.1 & 0.415 & 0.762 & 0.197\tabularnewline
\hline 
gpt-neox-20b & 0.507 & 0.603 & 0.041\tabularnewline
\hline 
flan-ul2 & \bf{0.560} & 0.747 & \bf{0.258}\tabularnewline
\hline
\end{tabular}
\end{table*}

\begin{table*}
\scriptsize
\caption{Cyber Security Benchmark Evaluation Results per Task.}
\label{tab: cyber-results}
    \centering
\begin{tabular}{|p{0.15\textwidth}|p{0.08\textwidth}|p{0.08\textwidth}|p{0.08\textwidth}|p{0.08\textwidth}|p{0.08\textwidth}|p{0.08\textwidth}|}
\hline 
Task &\multicolumn{5}{|c|}{Classification} & \multicolumn{1}{c|}{Summarization}\tabularnewline
\hline 
 & SPEC5G & CTI-to-MITRE with NLP & TRAM & SecureNLP & IoTSpotter & SPEC5G Summarization \tabularnewline
\hline 
\hline 
Metrics & Weighted F1 & F1 & Macro F1 & Binary F1 & Binary F1 & Rouge-L\tabularnewline
\hline
phi-3-5-mini-instruct & \bf{0.527} & 0.801 & 0.532 & 0.328 & 0.814 & 0.179\tabularnewline
\hline
mistral-7b-instruct-v0-3 & 0.517 & 0.798 & 0.532 & 0.283 & 0.812 & 0.187\tabularnewline
\hline 
llama2.7b & 0.432 & 0.541 & 0.530 & 0.265 & 0.601 & 0.450\tabularnewline
\hline 
llama2.7b.chat & 0.526 & 0.519 & 0.441 & 0.295 & 0.828 & 0.368\tabularnewline
\hline 
llama2.13b & 0.359 & 0.588 & \bf{0.554} & 0.284 & 0.689 & 0.444\tabularnewline
\hline 
llama2.13b.chat & 0.468 & 0.625 & 0.500 & 0.291 & 0.829 & 0.408\tabularnewline
\hline 
llama2.70b & 0.515 & 0.794 & 0.534 & 0.361 & \bf{0.905} & 0.411\tabularnewline
\hline 
llama2.70b.chat & 0.446 & \bf{0.812} & 0.377 & 0.303 & 0.874 & 0.365\tabularnewline
\hline 
granite.13b.v2 (base) & 0.193 & 0.430 & 0.190 & 0.287 & 0.033 & 0.397\tabularnewline
\hline 
granite.13b.instruct.v2 & 0.517 & 0.702 & 0.227 & 0.333 & 0.671 & 0.342\tabularnewline
\hline 
granite.13b.chat.v2.1 & 0.513 & 0.686 & 0.428 & 0.293 & 0.898 & 0.416\tabularnewline
\hline 
gpt-neox-20b & 0.163  & 0.128 & 0.111 & 0.021 & 0.000 & \bf{0.453}\tabularnewline
\hline 
flan-ul2 & 0.077 & 0.764 & 0.349 & \bf{0.369} & 0.869 & 0.331 \tabularnewline
\hline
\end{tabular}
\end{table*}


\subsection{Prompts} \label{sec:appendix-prompts}

\begin{figure*}[htb]
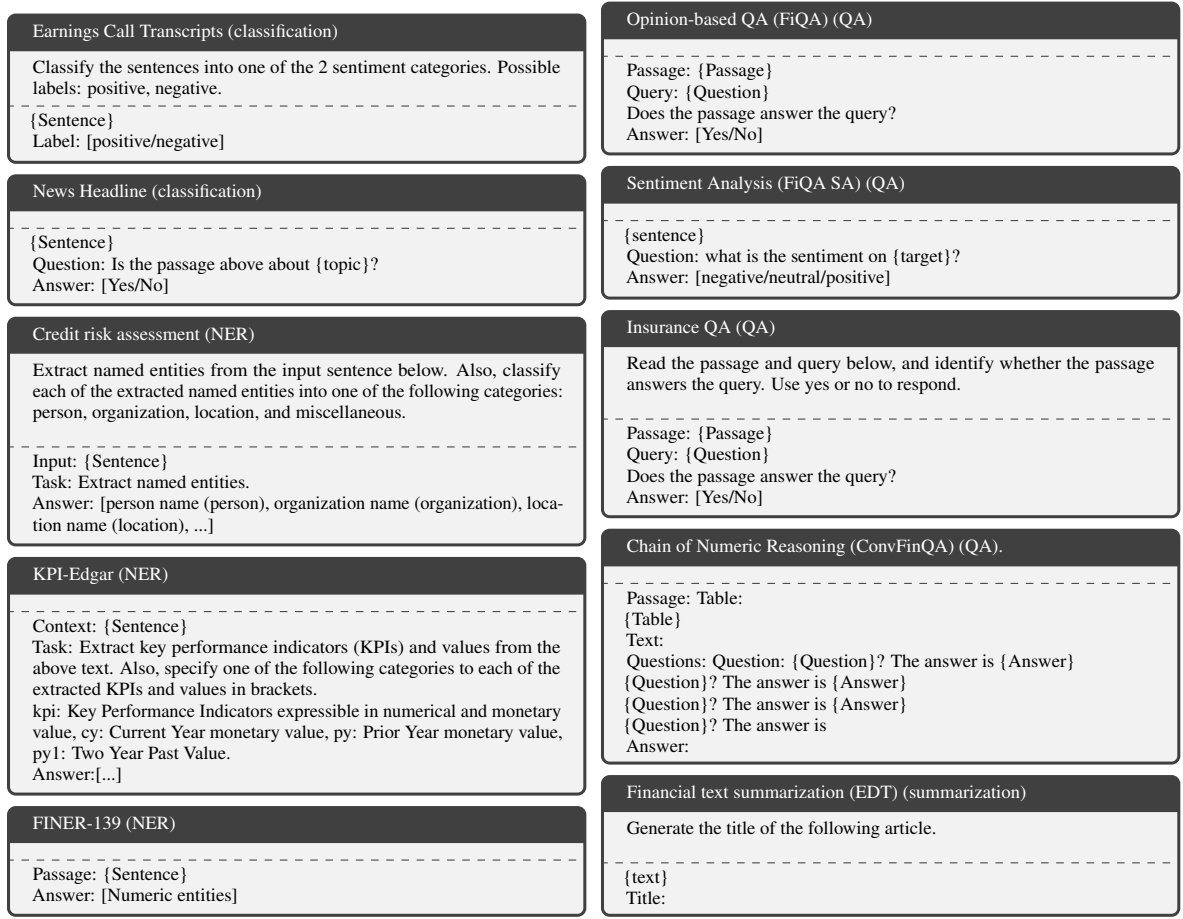


\begin{subfigure}{.99\columnwidth}
\begin{promptbox}{Earnings Call Transcripts (classification)}
Classify the sentences into one of the 2 sentiment categories. Possible labels: positive, negative.
\tcblower
\{Sentence\}\\
Label: [positive/negative]
\end{promptbox}
\begin{promptbox}{News Headline (classification)}
\tcblower
\{Sentence\}\\
Question: Is the passage above about \{topic\}?\\
Answer: [Yes/No]
\end{promptbox}
\begin{promptbox}{Credit risk assessment (NER)}
Extract named entities from the input sentence below. Also, classify each of the extracted named entities into one of the following categories: person, organization, location, and miscellaneous.\\
\tcblower
Input: \{Sentence\}\\
Task: Extract named entities.\\
Answer: [person name (person), organization name (organization), location name (location), ...]
\end{promptbox}
\begin{promptbox}{KPI-Edgar (NER)}
\tcblower
Context: \{Sentence\}\\
Task: Extract key performance indicators (KPIs) and values from the above text. Also, specify one of the following categories to each of the extracted KPIs and values in brackets.\\
kpi: Key Performance Indicators expressible in numerical and monetary value, cy: Current Year monetary value, py: Prior Year monetary value, py1: Two Year Past Value.\\
Answer:[...]
\end{promptbox}
\begin{promptbox}{FINER-139 (NER)}
\tcblower
Passage: \{Sentence\}\\
Answer: [Numeric entities]
\end{promptbox}
\end{subfigure}\hspace{0.5em}%
\begin{subfigure}{.99\columnwidth}
\begin{promptbox}{Opinion-based QA (FiQA) (QA)}
\tcblower
Passage: \{Passage\}\\
Query: \{Question\}\\
Does the passage answer the query?\\
Answer: [Yes/No]
\end{promptbox}
\begin{promptbox}{Sentiment Analysis (FiQA SA) (QA)}
\tcblower
\{sentence\}\\
Question: what is the sentiment on \{target\}?\\
Answer: [negative/neutral/positive]
\end{promptbox}
\begin{promptbox}{Insurance QA (QA)}
Read the passage and query below, and identify whether the passage answers the query. Use yes or no to respond.\\
\tcblower
Passage: \{Passage\}\\
Query: \{Question\}\\
Does the passage answer the query?\\
Answer: [Yes/No]
\end{promptbox}
\begin{promptbox}{Chain of Numeric Reasoning (ConvFinQA) (QA). }
\tcblower
Passage: Table: \\
\{Table\}\\
Text: \\
Questions:  Question: \{Question\}? The answer is \{Answer\} \\
\{Question\}? The answer is \{Answer\}\\
\{Question\}? The answer is \{Answer\}\\
\{Question\}? The answer is \\
Answer:
\end{promptbox}
\begin{promptbox}{Financial text summarization (EDT) (summarization)}
Generate the title of the following article.\\

\tcblower
\{text\}\\
Title:
\end{promptbox}
\end{subfigure}

\caption{Prompts of finance scenarios.}
\label{fig:prompts-finance}

\end{figure*}
\begin{figure*}[htb]
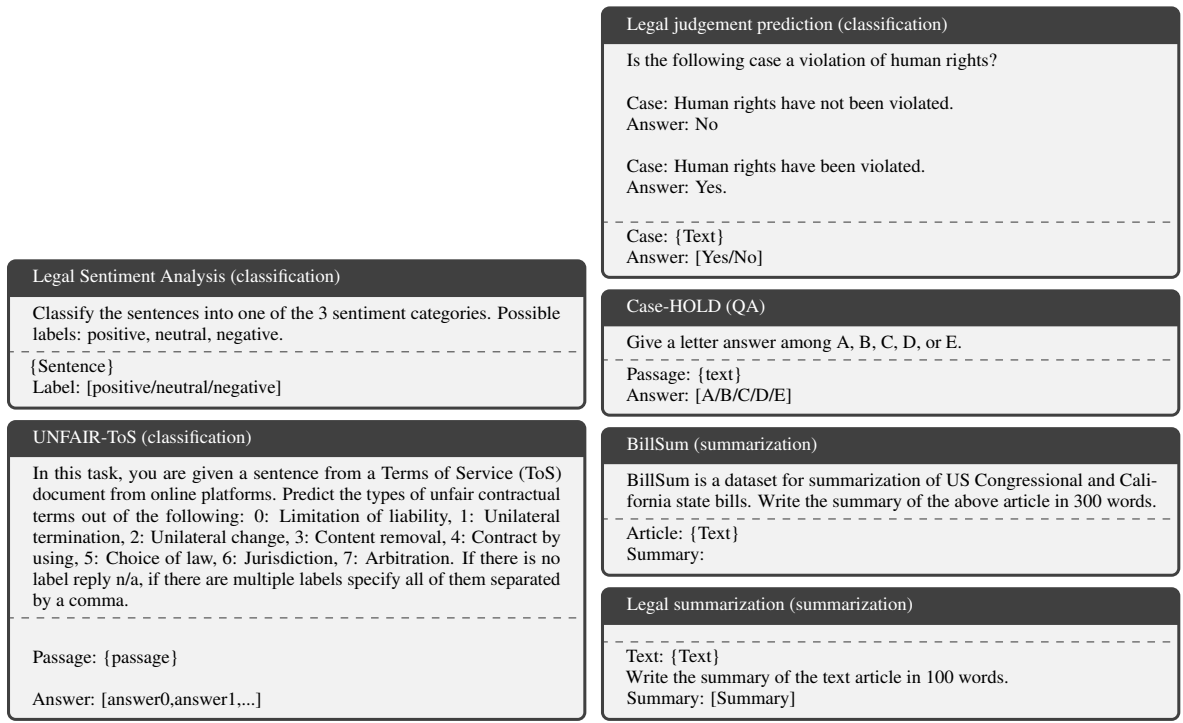

\begin{subfigure}{.99\columnwidth}
\begin{promptbox}{Legal Sentiment Analysis (classification)}
Classify the sentences into one of the 3 sentiment categories. Possible labels: positive, neutral, negative.
\tcblower
\{Sentence\}\\
Label: [positive/neutral/negative]
\end{promptbox}
\begin{promptbox}{UNFAIR-ToS (classification)}
In this task, you are given a sentence from a Terms of Service (ToS) document from online platforms. Predict the types of unfair contractual terms out of the following: 
0: Limitation of liability, 
1: Unilateral termination, 
2: Unilateral change, 
3: Content removal, 
4: Contract by using, 
5: Choice of law, 
6: Jurisdiction, 
7: Arbitration. 
If there is no label reply n/a, if there are multiple labels specify all of them separated by a comma.
\tcblower
\vspace{3ex}
Passage: \{passage\}\\
\\
Answer: [answer0,answer1,...]
\end{promptbox}
\end{subfigure}\hspace{0.5em}
\begin{subfigure}{.99\columnwidth}
\begin{promptbox}{Legal judgement prediction (classification)}
Is the following case a violation of human rights?\\
\\
Case: Human rights have not been violated.\\
Answer: No\\
\\
Case: Human rights have been violated.\\
Answer: Yes.\\
\tcblower
Case: \{Text\}\\
Answer: [Yes/No]
\end{promptbox}

\begin{promptbox}{Case-HOLD (QA)}
Give a letter answer among A, B, C, D, or E.
\tcblower
Passage: \{text\}\\
Answer: [A/B/C/D/E]
\end{promptbox}
\begin{promptbox}{BillSum (summarization)}
BillSum is a dataset for summarization of US Congressional and California state bills.
Write the summary of the above article in 300 words.
\tcblower
Article: \{Text\}\\
Summary:
\end{promptbox}
\begin{promptbox}{Legal summarization (summarization)}
\tcblower
Text: \{Text\}\\
Write the summary of the text article in 100 words.\\
Summary: [Summary]
\end{promptbox}
\end{subfigure}

\caption{Prompts of legal scenarios.}
\label{fig:prompts-legal}

\end{figure*}
\begin{figure*}[htb]
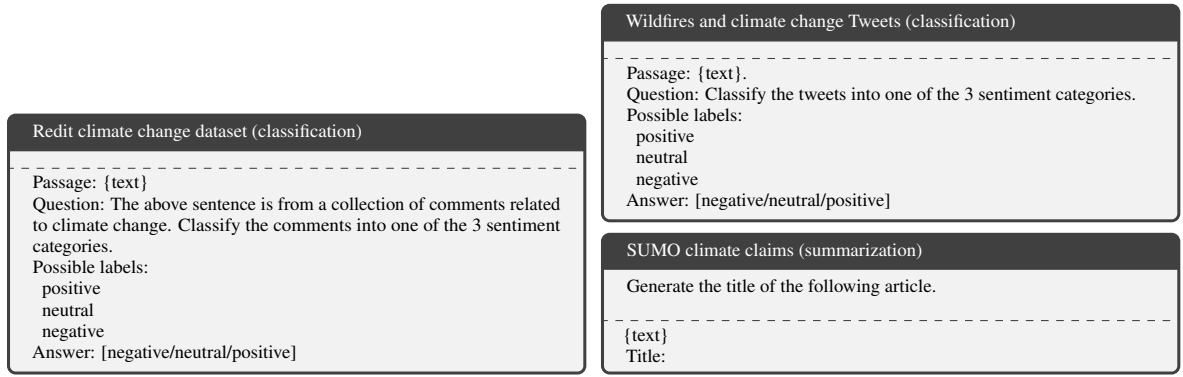

\begin{subfigure}{.99\columnwidth}
\centering
\begin{promptbox}{Redit climate change dataset (classification)}
\tcblower
Passage: \{text\}                 \\
Question: The above sentence is from a collection of comments related to climate change. Classify the comments into one of the 3 sentiment categories. \\
Possible labels:\\
\hspace*{0.5em}positive\\
\hspace*{0.5em}neutral\\
\hspace*{0.5em}negative\\
Answer: [negative/neutral/positive]
\end{promptbox}
\end{subfigure}\hspace{0.5em}
\begin{subfigure}{.99\columnwidth}
\begin{promptbox}{Wildfires and climate change Tweets (classification)}
\tcblower
Passage: \{text\}.                  \\
Question: Classify the tweets into one of the 3 sentiment categories. \\
Possible labels:\\
\hspace*{0.5em}positive\\
\hspace*{0.5em}neutral\\
\hspace*{0.5em}negative\\
Answer: [negative/neutral/positive]
\end{promptbox}
\begin{promptbox}{SUMO climate claims (summarization)}
Generate the title of the following article.\\
\tcblower
\{text\}\\
Title:
\end{promptbox}
\end{subfigure}

\caption{Prompts of climate scenarios}
\label{fig:prompts-climate}
\end{figure*}
\begin{figure*}[h]
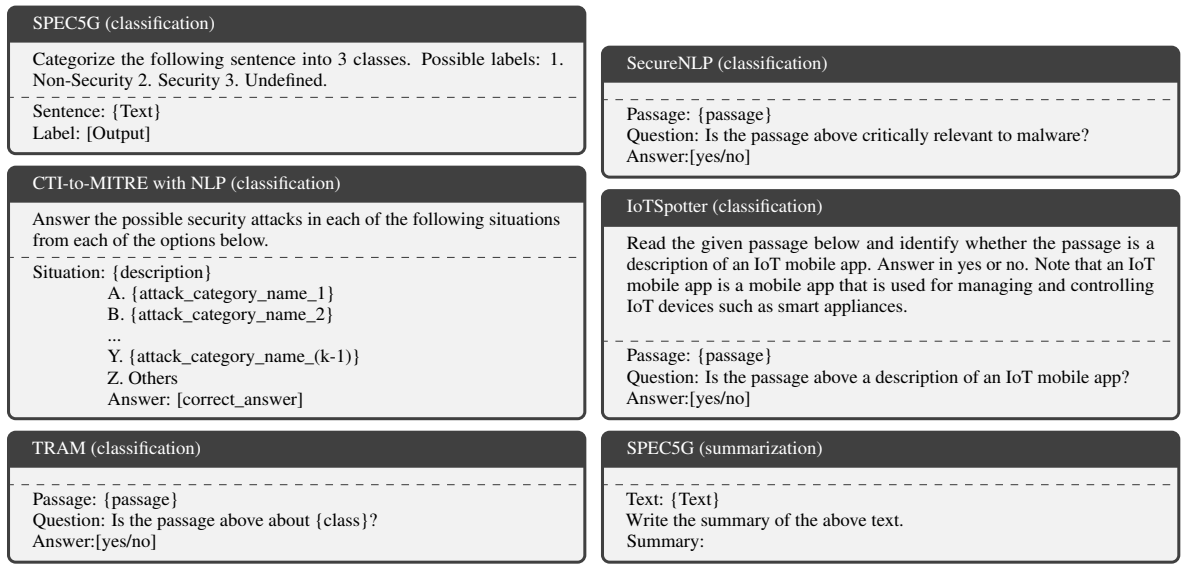

\begin{subfigure}{.99\columnwidth}
\centering
\begin{promptbox}{SPEC5G (classification)}
Categorize the following sentence into 3 classes. Possible labels: 1. Non-Security 2. Security 3. Undefined.
\tcblower
Sentence: \{Text\}\\
Label: [Output]
\end{promptbox}
\begin{promptbox}{CTI-to-MITRE with NLP (classification)}
Answer the possible security attacks in each of the following situations from each of the options below.                                   
\tcblower
Situation: \{description\}\hspace*{4.0em}\\
\hspace*{4.0em}A. \{attack\_category\_name\_1\}\\
\hspace*{4.0em}B. \{attack\_category\_name\_2\}\\
\hspace*{4.0em}...\\
\hspace*{4.0em}Y. \{attack\_category\_name\_(k-1)\}\\
\hspace*{4.0em}Z. Others\\
\hspace*{4.0em}Answer: [correct\_answer]
\end{promptbox}
\begin{promptbox}{TRAM (classification)}
\tcblower
Passage: \{passage\}\\
Question: Is the passage above about \{class\}?\\
Answer:[yes/no]
\end{promptbox}
\end{subfigure}\hspace{0.5em}
\begin{subfigure}{.99\columnwidth}
\begin{promptbox}{SecureNLP (classification)}
\tcblower
Passage: \{passage\}\\
Question: Is the passage above critically relevant to malware?\\
Answer:[yes/no]
\end{promptbox}
\begin{promptbox}{IoTSpotter (classification)}
Read the given passage below and identify whether the passage is a description of an IoT mobile app. Answer in yes or no. Note that an IoT mobile app is a mobile app that is used for managing and controlling IoT devices such as smart appliances.\\
\tcblower
Passage: \{passage\}\\
Question: Is the passage above a description of an IoT mobile app?\\
Answer:[yes/no]
\end{promptbox}
\begin{promptbox}{SPEC5G (summarization)}
\tcblower
Text: \{Text\}\\
Write the summary of the above text.\\
Summary:
\end{promptbox}
\end{subfigure}

\caption{Prompts for cybersecurity scenarios.}
\label{fig:prompts-cybersecurity}

\end{figure*}

Prompts that are used in the experiments are shown in this section.
Figures \ref{fig:prompts-finance} to \ref{fig:prompts-cybersecurity} shows the prompts for finance, legal, climate and sustainability, and cybersecurity scenarios, respectively.

A prompt consists of an "instruction" block, which is shown above a dotted line,
and an "input-output" block, which is shown below the dotted line. 
The instruction block contains an instruction,
which is placed at the beginning of a prompt. Some scenarios may not have the instruction block.
The input-output block contains a pair of the input and output 
of each sample. This is located after the instruction block. Within a block, a text enclosed with curly brackets \{ ... \} is replaced with an input text
of each sample. A text enclosed with square brackets [ ... ] is a placeholder of the generated text by
an LLM as an output. In the case of a few-shot learning setting, 
the input-output block can be used to show a training example for in-context learning. In that case, the placeholder of the output
is filled with the ground truth label of the sample. 
Such instances of input-output blocks that correspond to the few-shot examples are
iterated after the instruction block for $n$ times, where $n$ is the number of the shots of the in-context learning.
After the in-context learning examples, another input-output block is placed without filling the output with a ground truth label.




\end{document}